\documentclass[sigconf, screen]{acmart}

\makeatletter
\renewcommand\@formatdoi[1]{\ignorespaces}
\makeatother

\usepackage{multirow}
\usepackage{amsmath}
\usepackage[inline]{enumitem} 
\usepackage{xcolor}

\usepackage{hyperref}
\usepackage{lipsum}

\usepackage{arydshln}
\makeatletter
\def\adl@drawiv#1#2#3{
        \hskip.5\tabcolsep
        \xleaders#3{#2.5\@tempdimb #1{1}#2.5\@tempdimb}%
                #2\z@ plus1fil minus1fil\relax
        \hskip.5\tabcolsep}
\newcommand{\cdashlinelr}[1]{%
  \noalign{\vskip\aboverulesep
          \global\let\@dashdrawstore\adl@draw
          \global\let\adl@draw\adl@drawiv}
  \cdashline{#1}
  \noalign{\global\let\adl@draw\@dashdrawstore
          \vskip\belowrulesep}}
\makeatother

\copyrightyear{2023} 
\acmYear{2023} 
\setcopyright{none}
\acmConference[WWW '23]{Proceedings of the ACM Web Conference}{April 30--May 4, 2023}{Austin, TX, USA}
\acmBooktitle{Proceedings of the ACM Web Conference (WWW '23), April 30--May 4, 2023, Austin, TX, USA}

\begin{document}
\title{Practical Bandits: An Industry Perspective} 

\author{Bram van den Akker}
\affiliation{
  \institution{Booking.com}
  \city{Amsterdam}
  \country{The Netherlands}
}

\author{Olivier Jeunen}
\affiliation{
  \institution{ShareChat}
  \city{Edinburgh}
  \country{United Kingdom}
}

\author{Ying Li}
\affiliation{
  \institution{Netflix}
  \city{Los Gatos}
  \country{CA, USA}
}

\author{Ben London}
\affiliation{
  \institution{Amazon}
  \city{Seattle}
  \country{WA, USA}
}

\author{Zahra Nazari}
\affiliation{
  \institution{Spotify}
  \city{New York}
  \country{NY, USA}
}

\author{Devesh Parekh}
\affiliation{
  \institution{Netflix}
  \city{Los Gatos}
  \country{CA, USA}
}

\begin{abstract}
The bandit paradigm provides a unified modeling framework for problems that require decision-making under uncertainty.
Because many business metrics can be viewed as \emph{rewards} (a.k.a. \emph{utilities}) that result from \emph{actions}, bandit algorithms have seen a large and growing interest from industrial applications, such as search, recommendation and advertising.
Indeed, with the bandit lens comes the promise of direct optimisation for the metrics we care about.

Nevertheless, the road to successfully applying bandits in production is not an easy one.
Even when the action space and rewards are well-defined, practitioners still need to make decisions regarding multi-arm or contextual approaches, on- or off-policy setups, delayed or immediate feedback, myopic or long-term optimisation, etc.
To make matters worse, industrial platforms typically give rise to large action spaces in which existing approaches tend to break down.
The research literature on these topics is broad and vast, but this can overwhelm practitioners, whose primary aim is to solve practical problems, and therefore need to decide on a specific instantiation or approach for each project.
This tutorial will take a step towards filling that gap between the theory and practice of bandits.
Our goal is to present a unified overview of the field and its existing terminology, concepts and algorithms---with a focus on problems relevant to industry.
We hope our industrial perspective will help future practitioners who wish to leverage the bandit paradigm for their application.
\end{abstract}

%
%



\maketitle

\section{Introduction \& Motivation}
Modern-day platforms on the web consist of many moving parts, all making many decisions over time.
These decisions can include ranking of content on a homepage, recommendations, notifications and pricing---all personalised to the user being targeted.
Because of the complexity of such systems, manually optimising these decisions to achieve a common goal (such as retention, revenue, user satisfaction, and others) quickly becomes an unmanageable task.
For this reason, many businesses have adopted data-driven approaches to decision-making, so as to scale up while optimising for the desired success metrics.

As a research problem, sequential algorithmic decision-making under uncertainty predates modern web technology, with some of its earliest formulations being credited to \citeauthor{Robbins1952}~\cite{Robbins1952}.
In essence, a decision-maker needs to iteratively take an \emph{action} from a certain action space for a number of \emph{rounds}, with the goal of maximising the cumulative \emph{rewards} (a.k.a. \emph{returns} or \emph{utilities}) that the decision-maker obtains.
This seminal work has inspired a rich body of literature focusing on all aspects of what is now known as ``the bandit problem,''\footnote{This nomenclature originates from an analogy with a gambler who aims to maximise winnings across slot machines, often referred to as ``one-armed bandits.''} ultimately laying the groundwork for recent impressive advances in reinforcement learning~\cite{Silver18,Vinyals19}.
As such, there is promise in adopting the bandit perspective for problems faced by the online businesses.

There is an abundance of high quality literature introducing researchers and practitioners to the theory behind the bandit paradigm and its algorithms \cite{Slivkins19, lattimore2020bandit, Sutton1998}.
However, identifying work relevant to the everyday challenges faced by industry practitioners can be cumbersome. 
Indeed, algorithmic advances that lead to empirical progress typically stem from fundamental theoretical groundwork, which can be hard to navigate for those who wish to use these algorithms to their fullest potential, whilst remaining focused on practice.

It is our goal to take a step towards bridging this gap between the theory and practice of bandits.
This tutorial is the product of an industry-wide collaboration, giving rise to a diverse industrial perspective of using bandits in practice, for a wide range of applications.


\section{Proposal Details}

\subsection{Topic \& Relevance}

All authors have first-hand experience using bandit algorithms for various applications in large-scale web platforms, such as e-commerce and content streaming.

We have curated a family of topics based on the challenges practitioners face.
Each topic has seen a large number of recent contributions from both academia and industry, which can feel intimidating to navigate for those with limited experience or exposure to the bandit paradigm and its many facets.

\subsubsection{The Bandit Setting}
A first step in navigating the appropriate literature is to properly characterise the problem setting~\cite{Liu2021_marble}.

In its most general form, a bandit setting arises when learning without full-information (i.e., partial supervision)~\cite{cesa2006prediction}.
The appropriate algorithms vary when shifting between a \emph{full-bandit} setting, where only one of the actions is taken at the time, or a form of \emph{semi-bandits} where this is extended to multiple actions~\cite{lattimore2020bandit}.
Actions can be combinatorial, ranking, discrete or continuous; rewards can be single- or multi-objective, delayed or immediate.
We aim to provide a general overview, clarifying concepts and guiding practitioners to the relevant literature for their problem. 
As much as possible, we will ground each bandit formalism in a real application (e.g., by mapping ranking bandits to a recommender system example).

Note that bandits can be seen as a special case of Reinforcement Learning (RL) where state space is restricted to a single state~\cite{Sutton1998}.
While many bandit settings can naturally be extended into a more general RL formulation, we will not venture into this area of research too deeply.
However, we will make it clear when the limitations of the bandit formalism can potentially be overcome with RL, and provide pointers to relevant literature when applicable.

\subsubsection{On-policy}
The canonical bandit setting is the online, \emph{on-policy} setting, wherein an agent learns a policy by interacting directly with the environment in a sequence of decisions.
This gives rise to the well-known \emph{explore}-\emph{exploit} dilemma, in which the policy must weigh the informational value of taking an action with the reward it will yield.
Seminal practical applications of such methods in web platforms make use of Upper-Confidence-Bounds (UCB)~\cite{Li2010} or Thompson Sampling (TS)~\cite{Chapelle2011}.
These strategies can provably attain optimal regret, which intuitively makes them attractive~\cite{Auer2002}.
Nevertheless, elegant theoretical results often rely on overly restrictive assumptions---about the stationarity of the environment, and the policy's ability to \emph{immediately} observe the outcomes and update parameters---which rarely translate to real applications.
Indeed, deploying such real-time updates at scale is far from trivial, and it typically ignores that an on-policy bandit is often only part of a larger dynamic system.
As such, while on-policy algorithms are well-studied in the literature, and our understanding of them contributes to advances in related settings, examples of \emph{off-policy} methods being used in practice are much more abundant.






\subsubsection{Off-policy}
In practice, on-policy algorithms can be expensive to run and difficult to make tractable.
Furthermore, such instantly self-updating methods can give rise to safety and fairness concerns when they cannot be monitored appropriately~\cite{Jagerman2020}.

As such, there are many good reasons to move the model development stage off-line.
Here, we leverage logged data (\emph{contexts}, \emph{actions} and \emph{rewards}) from an existing policy in production (often referred to as the \emph{logging} policy), to learn~\cite{Swaminathan2015, Joachims2018, wang2016learning} and evaluate~\cite{saito2021counterfactual, swaminathan2017off} new policies,
This type of setting is called the ``off-policy'' setting --- and when the logging policy differs significantly from the policies we are trying to evaluate, this task becomes challenging.
Indeed, acess to logged data suitable for off-policy learning is not guaranteed~\cite{wang2016learning, joachims2017unbiased, london2022control}.
Designing a good strategy for collecting counterfactuals through randomisation~\cite{cesa2017boltzmann} while balancing exploration~\cite{schnabel2018short} is both important and non-trivial.



%

\subsubsection{Pure Exploration}
Pure exploration is a family of methods that use the bandit paradigm to extend traditional A/B testing. The goal is to identify the best arm~\cite{garivier2016optimal} or an approximation~\cite{jamieson2014lil} while minimising the exposure to each arm. We will briefly cover pure exploration as it is commonly applied in industry. 

\subsubsection{Rewards}
Theoretical works in the bandit literature commonly assume rewards to be predefined and given.
However, the design of good rewards is both essential and non-trivial when working with Bandits in practise. Designing a reward is subject to balancing long and short-term metrics ~\cite{wu2017returning}, as well as competing metrics coming from different stakeholders~\cite{mehrotra2020bandit, drugan2013designing}. Moreover, rewards are generally observed under some delay~\cite{Ktena2019,lancewicki2021stochastic} requiring practitioners to estimate and mitigate the corresponding biases. 

\subsubsection{Large Action Spaces}
Classical work in the bandit literature typically deals with relatively small action spaces --- especially when compared to recommendation setups where the cardinality of the item catalogue spans the order of millions.
Because bounds on regret typically depend on the size of the action space, this is to be expected.
Successful applications have been reported for bandit-based decision-making when the number of options is limited: for example, choosing which playlists from a predefined set to show on a user's homepage~\cite{Gruson2019,Bendada2020}.
Recent advances provide ways forward, where action features can be leveraged to transfer signal between arms, and possibly to new arms~\cite{Saito2022,London2020_NeurIPS}.
When the action space is continuous~\cite{Jeunen2022_AdKDD} or consists of rankings~\cite{sen2021top,Kiyohara2022}, we need to explicitly reason about the structure of the problem to allow for sample-efficient learning and estimation.

Additionally, approximate nearest-neighbour retrieval methods are often of paramount importance for systems to scale appropriately~\cite{Malkov2020}.
We aim to provide an overview of recent advances in this important area, broadening the scope of problems to which bandit methods can be applied effectively and efficiently.

\subsubsection{Practical Considerations}
When applying bandit algorithms at scale, issues arise that are not always covered by the academic literature. Approximations must be made in order to meet strict latency demands. Online model updates typically happen in batches at a regular cadence, rather than immediately after each interaction. 
Moreover, in production systems that retrain and deploy new models without human intervention, automated validations are crucial. 
We can evaluate the ``goodness" of a policy using logged data, but these evaluations critically depend on the propensities of the data collection policy---which are not always available or reliable, and therefore must be approximated.
We will discuss these practical issues, the measures taken to address them and their repercussions.



\subsection{Style \& Duration}
This will be a 3-hour hands-on tutorial.
First, we aim to cover the fundamentals and introduce the state-of-the-art in research for practical bandit applications.
In doing so, it is our hope to bridge the scientific research and applied practitioner communities interested in this field.
Second, we will dive deeper into specific applications of bandits in practice, outlining how theoretical approaches map to practical instantiations.
Third, we will cover hands-on Jupyter notebooks that provide insights into how these approaches can be implemented in the OpenBandit Pipeline~\cite{Saito2021}\footnote{\href{https://github.com/st-tech/zr-obp}{github.com/st-tech/zr-obp}}, and how one might deal with typical industrial challenges~\cite{VandenAkker2022}.
These notebooks will run on Google Colaboratory\footnote{\href{https://colab.research.google.com/}{colab.research.google.com}}, allowing attendees to run examples in their browsers without requiring the installation of any software.

We will build to advanced algorithms from the ground up, making this tutorial an adequate starting point for novice as well as experienced researchers and practitioners in this fast-growing field.

\subsection{Audience}
This tutorial targets every researcher and practitioner that has considered the use of bandit algorithms to solve a practical problem.
The material requires an expected background knowledge on a MSc. level in computer science, machine learning or related fields, and will build its way up to advanced algorithms from fundamentals.
Because of our practical focus, fundamental theory will be linked to practical considerations when applicable.
We expect attendees to feel empowered to devise practical bandit-based solutions to machine learning problems they will be faced with in the future.


\subsection{Previous Editions}
This will be the first edition of this tutorial, but the presenters have experience in teaching material covering similar topics in the past.
We anticipate to present (an extended version of) this tutorial at similar conferences in the future.

\subsection{Tutorial Material}
We will provide all the slides as well as a curated reading list covering relevant literature to participants.
The notebooks used for the hands-on session will additionally be made available, all hosted in a public GitHub repository.
If possible, we would like to record our tutorial and share the recordings with interested participants afterwards.

\subsection{Equipment}
No additional equipment will be necessary---but it would be preferable if recording equipment is available in the room.
Otherwise, we will use a recording setup with Zoom.

\subsection{Video Teaser}
The video teaser can be found at \href{https://youtu.be/lHva_kgRqq4}{https://youtu.be/lHva\_kgRqq4}.

\subsection{Organisation Details}
We will host all materials on a publicly available GitHub repository.
This includes the slides that will be covered during the tutorial, as well as the notebooks for the hands-on sessions.
Because the material will be self-contained and the notebooks will run on Google Colaboratory, this lends itself to an asynchronous setup where interested participants can go through the materials at their own pace at a later date.
Additionally, we wish to record the presentations so they can be distributed afterwards.


\section{Organiser Biographies}
The workshop organisers and their biographies are listed here in alphabetical order:

\begin{description}
    \item[Bram van den Akker] (\href{mailto:bram.vandenakker@booking.com}{bram.vandenakker@booking.com}) is a Senior Machine Learning Scientist at Booking.com with a background in Computer Science and Artificial Intelligence from the University of Amsterdam. At Booking.com, Bram's work focuses on bridging the gap between  applied research and practical requirements for Bandits all across the company. Previously, Bram has held positions at Shopify \& Panasonic, and has peer reviewed contributions to conferences and workshops such as TheWebConf, RecSys, and KDD of which one has been awarded with a best-paper award~\cite{van2019vitor}.
    
    \item[Olivier Jeunen] (\href{mailto:jeunen@sharechat.co}{jeunen@sharechat.co}) is a Lead Decision Scientist at ShareChat with a PhD from the University of Antwerp \cite{Jeunen2021thesis}, having previously held positions at Amazon, Spotify, Facebook and Criteo.
    Olivier's research focuses on applying ideas from causal and counterfactual inference to recommendation and advertising problems, which have led to 20+ peer reviewed contributions to top-tier journals, conferences, and workshops at NeurIPS, KDD, RecSys and WSDM, of which two have been recognised with best paper awards~\cite{Jeunen2021A,Jeunen2022_AdKDD}. He is an active PC member for KDD, RecSys, The WebConf, WSDM and SIGIR, whilst reviewing for several journals and workshops---which has led to two outstanding reviewer awards. He currently serves as co-Web Chair for RecSys, co-chaired the Dutch-Belgian Information Retrieval Workshop '20 and the CONSEQUENCES Workshop at RecSys '22~\cite{CONSEQUENCES22}, and co-lectured tutorials at the RecSys Summer School '19, UMAP '20~\cite{Vasile2020} and The WebConf '21~\cite{Vasile2021}.
    
    \item[Ying Li] (\href{mailto:yingl@netflix.com}{yingl@netflix.com}) is a Senior Research Scientist at Netflix. She obtained her Ph.D. from the University of California, Los Angeles, and B.S. from Peking University. At Netflix, her research interest focuses on large-scale search and recommendation systems, bandit and long-term user satisfaction optimization. Prior to Netflix, she was an Applied Scientist in Amazon, focusing on cold-start classification and large-scale extreme classification using NLP. She has co-chaired the REVEAL (Reinforcement learning-based recommender systems at scale) workshop at RecSys 2022 \cite{liaw2022reveal}.
    
    \item[Ben London] (\href{mailto:blondon@amazon.com}{blondon@amazon.com})
    is a Sr.\ Scientist at Amazon Music.
    He earned his Ph.D. in 2015 at the University of Maryland, where he was advised by Lise Getoor and worked closely with Ben Taskar and Bert Huang. His research investigates machine learning theory and algorithms, with a focus on generalization guarantees, structured prediction, recommendation, contextual bandits, and evaluation/learning with logged bandit feedback. His work has been published in JMLR, ICML, NeurIPS, AISTATS, UAI and RecSys. He was a co-organizer of the NeurIPS 2019 Workshop on ML with Guarantees, an area chair for ICML (2020, 2022) and NeurIPS (2020, 2021), a senior PC member for IJCAI, and has reviewed for numerous conferences and journals.

    \item[Zahra Nazari] (\href{mailto:zahran@spotify.com}{zahran@spotify.com})
    is a Sr. Scientist at Spotify working on the design and evaluation of recommender systems with a focus on sequential and long term optimization using reinforcement learning. Prior to Spotify, she had held positions at Google, Twitter and Framehawk and earned her Ph.D. at the University of Southern California with a focus on preference elicitation and human behaviour modelling in complex situations such as negotiations. She has published 20+ papers in top conferences such as IJCAI, AAMAS, SIGIR, WSDM and The Web Conference.  Zahra is serving as the program committee member in conferences such as SIGIR, The Web Conference, CIKM and has co-organized the workshop on Multi-Objective Recommender Systems in Recsys for two consecutive years (2021 and 2022).    
        
    \item[Devesh Parekh] (\href{mailto:dparekh@netflix.com}{dparekh@netflix.com})
    is a Staff Research Engineer at Netflix working on simplifying the deployment of contextual bandit solutions to new problems in member-facing applications. Devesh has previously worked on causal inference and contextual bandit problems in the areas of programmatic marketing, signup optimization, and member engagement messaging at Netflix.
     
\end{description}


\bibliographystyle{ACM-Reference-Format}
\bibliography{bibliography}


\end{document}